\documentclass[letterpaper, 10 pt, journal, twoside]{ieeetran}

\IEEEoverridecommandlockouts                              

\usepackage{graphics} 
\usepackage{epsfig} 
\usepackage{mathptmx} 
\usepackage{times} 
\usepackage{amsmath} 
\usepackage{amssymb}  

\usepackage{xifthen}
\usepackage{xparse}
\usepackage{subdepth}
\usepackage{leftidx}
\usepackage{bm}
\usepackage{amsmath}
\newcommand{\norm}[1]{\left\Vert#1\right\Vert}

\newcommand{\bbm}{\begin{bmatrix}}
	\newcommand{\ebm}{\end{bmatrix}}

\DeclareMathAlphabet{\mbf}{OT1}{ptm}{b}{n}
\newcommand{\mbs}[1]{{\bm{#1}}}

\newcommand{\mbsbar}[1]{{\overline{\boldsymbol{#1}}}}
\newcommand{\mbshat}[1]{{\hat{\boldsymbol{#1}}}}
\newcommand{\mbstilde}[1]{{\tilde{\boldsymbol{#1}}}}

\newcommand{\mbsdot}[1]{{\dot {\boldsymbol{#1}}}}

\newcommand{\mbfbar}[1]{{\overline{\mbf{#1}}}}
\newcommand{\mbfhat}[1]{{\hat{\mbf{#1}}}}
\newcommand{\mbftilde}[1]{{\tilde{\mbf{#1}}}}

\newcommand{\mbfdot}[1]{{\dot{\mbf{#1}}}}

\newcommand{\cframe}[1]{{\smash{\protect\underrightarrow{\mathcal{F}}_{#1}}}}

\DeclareMathAlphabet{\mathbfit}{OML}{cmm}{b}{it}
\newcommand{\homo}[1]{{\mathbfit{#1}}}

\newcommand{\mbfh}[1]{{\homo{#1}}}

\newcommand{\argmin}{\operatornamewithlimits{argmin}}

\newcommand{\trans}[3]{\leftidx{_{#1}}{\mbf t}{\IfValueTF{#2}{_{#2#3\hspace{2pt}}}{}}} 

\newcommand{\vel}[3]{\leftidx{_{#1}}{\mbf v}{\IfValueTF{#2}{_{#2#3\hspace{2pt}}}{}}} 
\newcommand{\veltilde}[3]{\leftidx{_{#1}}{\mbftilde v}{\IfValueTF{#2}{_{#2#3\hspace{2pt}}}{}}} 
\newcommand{\velbar}[3]{\leftidx{_{#1}}{\mbfbar v}{\IfValueTF{#2}{_{#2#3\hspace{2pt}}}{}}} 
\newcommand{\velhat}[3]{\leftidx{_{#1}}{\mbfhat v}{\IfValueTF{#2}{_{#2#3\hspace{2pt}}}{}}} 
\newcommand{\veldot}[3]{\leftidx{_{#1}}{\mbfdot v}{\IfValueTF{#2}{_{#2#3\hspace{2pt}}}{}}} 

\newcommand{\acc}[3]{\leftidx{_{#1}}{\mbf a}{\IfValueTF{#2}{_{#2#3\hspace{2pt}}}{}}} 
\newcommand{\acctilde}[3]{\leftidx{_{#1}}{\mbftilde a}{\IfValueTF{#2}{_{#2#3\hspace{2pt}}}{}}} 
\newcommand{\accbar}[3]{\leftidx{_{#1}}{\mbfbar a}{\IfValueTF{#2}{_{#2#3\hspace{2pt}}}{}}} 

\newcommand{\rotvel}[3]{\leftidx{_{#1}}{\mbs \omega}{\IfValueTF{#2}{_{#2#3\hspace{2pt}}}{}}} 
\newcommand{\rotveltilde}[3]{\leftidx{_{#1}}{\mbstilde \omega}{\IfValueTF{#2}{_{#2#3\hspace{2pt}}}{}}} 
\newcommand{\rotvelbar}[3]{\leftidx{_{#1}}{\mbsbar \omega}{\IfValueTF{#2}{_{#2#3\hspace{2pt}}}{}}} 
\newcommand{\rotvelhat}[3]{\leftidx{_{#1}}{\mbshat \omega}{\IfValueTF{#2}{_{#2#3\hspace{2pt}}}{}}} 
\newcommand{\rotveldot}[3]{\leftidx{_{#1}}{\mbsdot \omega}{\IfValueTF{#2}{_{#2#3\hspace{2pt}}}{}}} 

\newcommand{\C}[2]{\leftidx{}{\mbf C}{_{#1#2\hspace{2pt}}}} 
\newcommand{\T}[2]{\leftidx{}{\mbfh T}{_{#1#2\hspace{2pt}}}} 

\newcommand{\generalThree}[3]{\leftidx{_{#1}}{\mbf #2}{_{#3}}} 
\newcommand{\pixel}[1]{{\mbfh u}{_{#1}}} 
\usepackage{multirow}
\usepackage[caption=false]{subfig}
\usepackage{tikz}
\usepackage{hyperref}

\title{Deep Probabilistic Feature-metric Tracking}

\author{Binbin Xu, Andrew J. Davison, and Stefan Leutenegger
\thanks{Manuscript received: September, 1, 2020; Revised November, 6, 2020; Accepted November, 13, 2020. This paper was recommended for publication by Editor Sven Behnke upon evaluation of the Associate Editor and Reviewers' comments.} 
\thanks{
This research is supported by Imperial College London and the EPSRC grant Aerial ABM EP/N018494/1. Binbin Xu holds a China Scholarship Council-Imperial Scholarship. }
\thanks{	
The authors are with Department of Computing, Imperial College London, United Kingdom. Corresponding author emails:
 	{\tt\small \{b.xu17, a.davison, s.leutenegger\}@imperial.ac.uk}}
\thanks{	
 	The supplementary video can also be watched on: \url{https://youtu.be/6pMosl6ZAPE}.
 	}
\thanks{Digital Object Identifier (DOI): see top of this page.}
}

\begin{document}

\markboth{IEEE Robotics and Automation Letters. Preprint Version. Accepted November, 2020}
{Xu \MakeLowercase{\textit{et al.}}: Deep Probabilistic Feature-metric Tracking} 

\maketitle

\begin{abstract}

Dense image alignment from RGB-D images remains a critical issue for real-world applications, especially under challenging lighting conditions and in a wide baseline setting. In this paper, we propose a new framework to learn a pixel-wise deep feature map and a deep feature-metric uncertainty map predicted by a Convolutional Neural Network (CNN), which together formulate a deep probabilistic feature-metric residual of the two-view constraint that can be minimised using Gauss-Newton in a coarse-to-fine optimisation framework. Furthermore, our network predicts a deep initial pose for faster and more reliable convergence. The optimisation steps are differentiable and unrolled to train in an end-to-end fashion. Due to its probabilistic essence, our approach can easily couple with other residuals, where we show a combination with ICP. Experimental results demonstrate state-of-the-art performances on the TUM RGB-D dataset and the 3D rigid object tracking dataset. We further demonstrate our method's robustness and convergence qualitatively. 

\end{abstract}


\section{INTRODUCTION}
\IEEEPARstart{D}{ense}  
image alignment~\cite{Lucas:Kanade:IJCAI1981} using the photometric residual has been widely applied in 2D tracking~\cite{Shi:Tomasi:CVPR1994}, 3D object tracking~\cite{Xu:etal:ICRA2019}, optical flow~\cite{Horn:Schunck:AI1981}, and SLAM~\cite{Newcombe:etal:ICCV2011}. In visual SLAM, it leads to two types of estimator designs: sparse~\cite{Engel:etal:PAMI2017} and dense type~\cite{Newcombe:etal:ICCV2011}. There has been an argument that dense methods that utilise information from all image pixels should exhibit better performance in terms of robustness and accuracy. However, this is not necessarily the case in reality, as investigated in~\cite{Platinsky:etal:ICRA2017}, especially compared to performance achieved by systems using the indirect sparse residual formulation (reprojection error)~\cite{Mur-Artal:etal:TRO2017}. 

One reason is that lighting change and reflection in real scenes break the brightness constancy assumption~\cite{Horn:Schunck:AI1981} commonly used in dense image alignment. Thus the resulting dense photometric residual cannot be well explained by the Gaussian distribution assumed in the Gauss-Newton scheme, which is in contrast to reprojection error minimisation that may still work robustly as long as sparse feature matches can be established. Secondly, the photometric residual considers only very local color consistency, which requires a good initialisation close to the global minimum. This leads to a poorer estimation accuracy when the baseline gets larger. On the contrary, the keypoint reprojection residual models a global constraint using a sparse feature descriptor, leading to better convergence properties.

\begin{figure}[t]
    \centering
    \includegraphics[width=1.0\columnwidth]{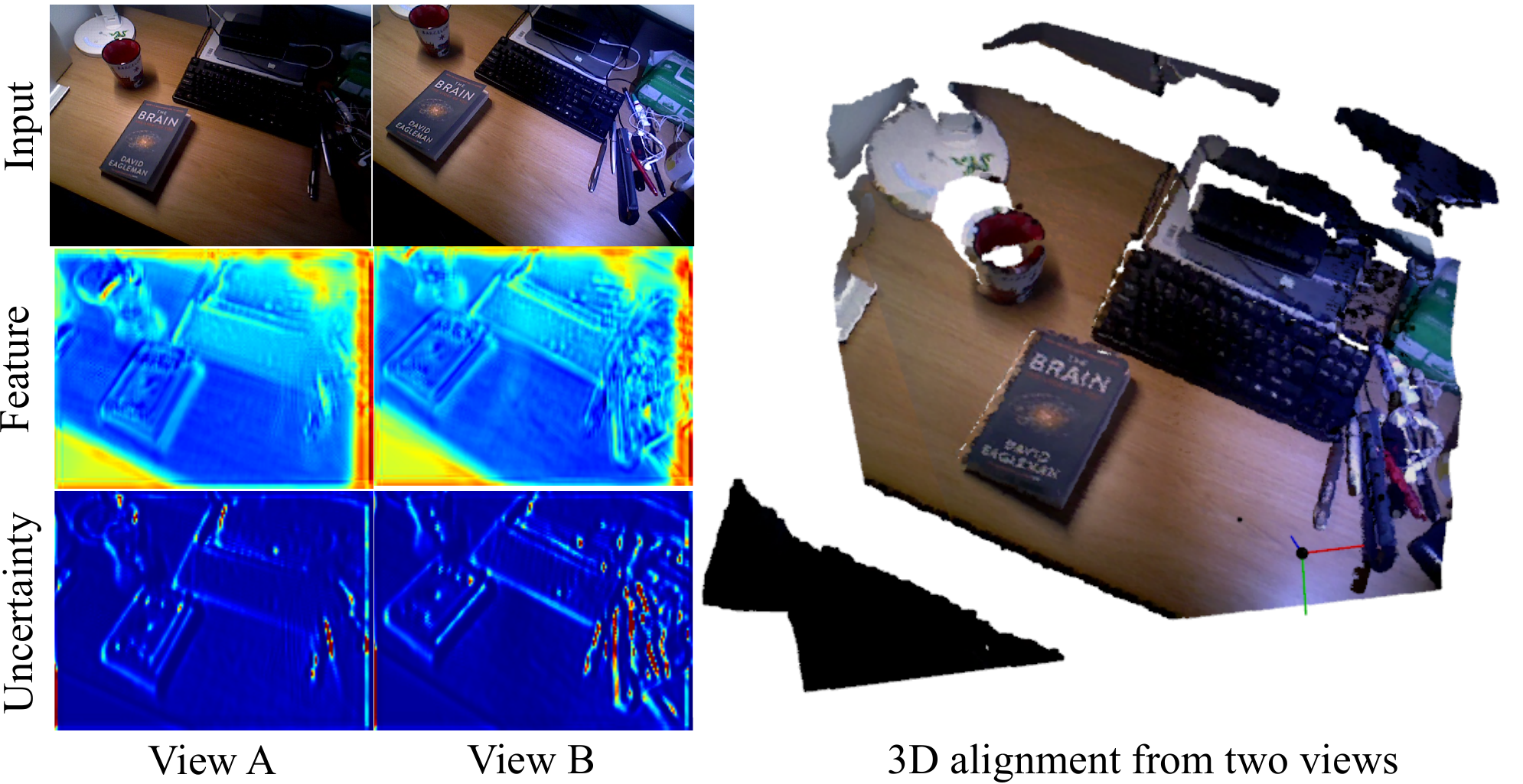}%
    \caption{We propose a probabilistic feature-metric tracking method that estimates dense feature and uncertainty maps from a pair of RGB-D images to optimise the relative pose between them. Our method can handle strong lighting changes and large motion scenarios by leveraging features that are robust to lighting changes, e.g. on the desk surface, and predicting high uncertainties on areas that the network cannot handle, e.g. for the strong lighting changes near the pens. }
    \label{fig:teaser}
\end{figure}

In this paper, we are trying to address these issues by replacing raw intensity image alignment with deep feature map alignment. Different from the existing learning-based feature-metric alignment~\cite{Czarnowski:etal:ICCVW2017,Lv:etal:CVPR2019, Tang:Tan:ICLR2019, Schmidt:etal:ICRA2017, Stumberg:etal:RAL2020}, we argue that the feature-metric residual should incorporate not simply the feature difference but also the corresponding uncertainty. Predictions from neural networks inherently are uncertain, which can be estimated~\cite{Kendall:Gal:NIPS2017}. Secondly, and also importantly, SLAM has most successfully been posed as a probabilistic problem, where uncertainty of the residuals has to be known~\cite{Thrun:etal:Book2005}, in particular when fusing different sensors and residuals. We will show how our feature-metric residuals can be combined with geometric ICP residuals using uncertainties to further improve results. The proposed probabilistic feature-metric residuals are minimised using coarse-to-fine Gauss-Newton optimisation. To ensure that the learned feature-metric cost landscape is suitable for the Gauss-Newton optimisation, we unroll the iterative optimisation steps and train the whole pipeline end-to-end. To handle the initialisation issue in the wide baseline case, we include training pairs with varied baselines and propose to replace the identity initialisation with a predicted initial pose from a pose network. This can improve the system convergence by bringing the initialisation into the convergence basin of the correct minimum. As shown in Fig.~\ref{fig:teaser}, the proposed method can handle large motion and strong illumination variance. The learned features are robust to lighting changes in most regions, e.g.\ reflection on desk surface, and the uncertainty map (red means high uncertainty) can downweigh the region, e.g.\ pens, where the feature predictions are uncertain.
In summary, we make the following contributions:
\begin{enumerate}
    \item We propose a dense probabilistic feature-metric residual, where a CNN predicts both feature and uncertainty maps used for non-linear least-squares minimisation to estimate the relative camera or object pose.
    \item In our CNN architecture, we propose a coupled feature encoder and pose predictor network, which combines the learning-based initial pose prediction and the learned features/uncertainties for pose optimisation, and train them together end-to-end.
    \item We further demonstrate how our proposed probabilistic feature-metric residual can easily lend itself to integration with other residuals, where a classic ICP residual is showcased. 
\end{enumerate}
We evaluate our proposed method on the TUM RGB-D SLAM dataset~\cite{Sturm:etal:IROS2012} and MovingObjects3D rigid motion dataset~\cite{Lv:etal:CVPR2019}. We provide ablation studies to validate each contribution component. We further provide a qualitative evaluation on the convergence basin and demonstrate the robustness under strong lighting changes.


\section{Related Work}
\textbf{Feature-metric Alignment:~}
To relax the brightness constancy constraint in direct image alignment, several recent works have exploited the feature-metric alignment by utilising features from neural networks. \cite{Jaramillo:etal:3DV2017, Czarnowski:etal:ICCVW2017} replace image intensity with 
high-dimensional features extracted from a pre-trained neural network for tracking and show a better robustness than using image intensity. However, the pre-trained features are not naturally consistent across different views and the redundancy in the pre-trained very high-dimensional features means a high cost of memory and computation time. 

\cite{Schmidt:etal:ICRA2017} proposes to learn a robust feature descriptor suitable for estimating dense correspondence in different lighting conditions and viewpoints using the contrastive loss~\cite{Hadsell:etal:CVPR2006}. \cite{Stumberg:etal:RAL2020} combines the contrastive loss with a Gauss-Newton loss, which includes a 2-dimensional pixel position uncertainty, to train dense
features. However, both of these works generate a feature map good for correspondence matching rather than alignment. The composed residuals do not necessarily fit well with the least square optimisation used for pose estimation. This is why \cite{Schmidt:etal:ICRA2017} requires a RANSAC step for refinement and \cite{Stumberg:etal:RAL2020} is only used for re-localisation. 

Recently, some methods start to explore how to combine the feature map learning more tightly with the least-square optimisation of camera tracking, based on the differentiable property of iterative optimisation. \cite{Wang:etal:ICRA2018} learn feature maps for 2D image tracking in the Lucas-Kanade framework. \cite{Tang:Tan:ICLR2019} propose feature-metric bundle adjustment for 3D reconstruction. \cite{Bloesch:etal:ICCV2019} propose to use feature maps for depth prediction and pose estimation. However, these works only consider a spatial correlation in feature generation, ignoring the temporal correlation in input image pairs. Quite related to our work, \cite{Lv:etal:CVPR2019} propose a spatio-temporal feature encoder by concatenating two views for the network input and further propose an m-estimator network and damping network for pose optimisation. However, different from ours, none of these works exploit feature-metric uncertainty in their settings, nor combine a pose predictor to boost convergence. 
\\\textbf{Deep Pose Prediction:~}
A different way to estimate pose from a pair of images is to leverage CNN predictions directly~\cite{Zhou:etal:CVPR2017, Ummenhofer:etal:ARXIV2016}. Learning a direct mapping from input images to 6D relative pose skips potential convergence issues of least-squares optimisation. However, it requires a large number of model parameters and a vast amount of training data, while not necessarily generalising to new scenes. 

To improve accuracy and generalisation, some recent works include coarse-to-fine estimation~\cite{Zhou:etal:ECCV2018} and iterative refinement~\cite{Li:etal:ECCV2018} to estimate a relative transformation. Despite some shared weights in iterations, these works still come with a much larger model capacity (i.e.\ parameter number) than the ones using optimisation -- even those with learned features -- and do not necessarily show an advantage in terms of pose accuracy. To better leverage both types of approaches, we propose a coarse-to-fine optimisation using learned features and uncertainties, plus a direct pose prediction on the coarsest layer serving as an \emph{initial guess}, which takes the output from the coarsest level two-view encoder as an input to make it compact.
\\\textbf{Uncertainty Learning:~}
Safety considerations have prompted recent works on uncertainty estimation of deep learning, as discussed in \cite{Kendall:Gal:NIPS2017} and applied to several tasks~\cite{Kendall:etal:CVPR2018}. \cite{Tateno:etal:CVPR2017} fuse the predicted depth into a monocular SLAM system and estimate the depth uncertainties via its difference with the nearest key-frame. \cite{Zhou:etal:ECCV2018} propose to estimate both depth and pose uncertainty in their depth and pose prediction networks. \cite{Liu:etal:CVPR2019} formulate the depth uncertainty differently using a probability volume.  Recently, D3VO~\cite{Yang:etal:CVPR2020} propose to estimate the photometric uncertainties and  predict a relative pose to initialise the pose optimisation. Most of these works, if not all, model the uncertainty based on the difference between the prediction and the ground truth values. In contrast to these works, we propose a novel feature-metric uncertainty and learn it without ground truth feature maps available in the training. Instead, we formulate the uncertainty in a novel probabilistic feature-metric residual and learn it implicitly as part of the least-squares optimisation. The learned features and uncertainties should lead to a better optimised pose via training back-propagation.    


\section{Method}
\begin{figure*}[htb]
    \centering
    \includegraphics[width=0.8\textwidth]{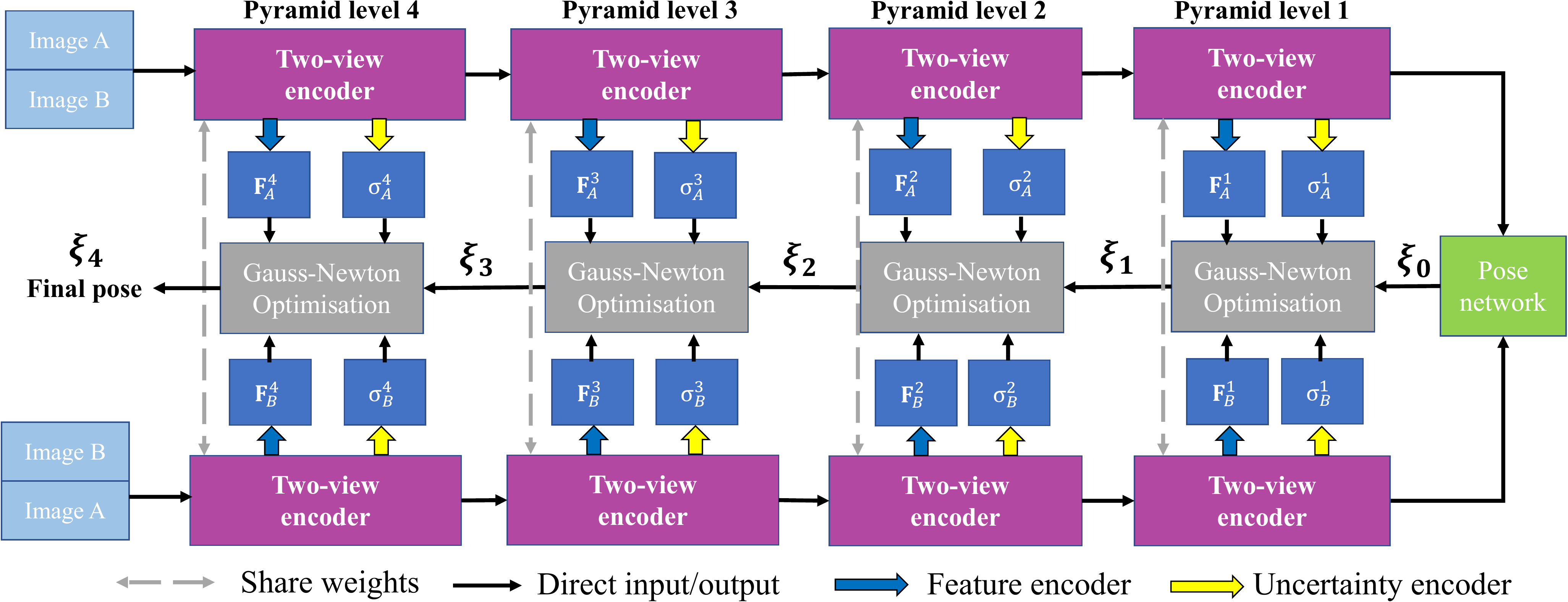}%
    \caption{Overview of our proposed deep probabilistic feature-metric tracking method. For two views, we input image $A$ and image $B$, by concatenating them as \{$A$, $B$\} and \{$B$, $A$\}, respectively, to our two-view encoder pyramid network. At each pyramid level, we extract the output from the two-view encoder and feed it into the feature encoder and uncertainty encoder separately to extract dense feature and uncertainty maps. Then we optimise the pose by minimising the proposed probabilistic feature-metric residual, which is initialised by the pose from the coarser level. On the coarsest level, we concatenate the outputs of the two views from the two frames and run through the pose network to obtain an initial pose prediction. 
    }
    \label{fig:pipeline}
\end{figure*}

Fig.~\ref{fig:pipeline} shows an overview of our system. For a pair of RGB-D frames, frame $A$ $\cframe{A}$ and frame $B$ $\cframe{B}$, our aim is to estimate its relative transformation $\T{A}{B}=(\C{A}{B},\trans{A}{A}{B}) \in (SO(3) \times \mathbf{R}^3)$, from $\cframe{B}$ to $\cframe{A}$. We represent $\T{A}{B}$ in twist coordinates $\mbs \xi$ by $\T{A}{B} \left(\mbs \xi\right)=\exp(\mbs \xi_{AB})$. Each frame has a depth map $\textbf{D}$ and a color image $\textbf{I}$. The network components in our whole system are denoted as $\phi$,
with the two-view spatio-temporal encoder $\phi_\theta$, the feature encoder $\phi_F$, the uncertainty encoder $\phi_\sigma$, and the pose network $\phi_T$. The weights are shared across the two views for $\phi_\theta$, $\phi_F$, and $\phi_\sigma$. The architecture details of all our network components can be found in the subsection~\ref{subsec: implementation details}.

To extract the spatial and temporal correlation between two frames, we first concatenate the input colour and depth image along the feature channel and feed them through the two-view spatio-temporal encoder pyramid network: 
\begin{equation}
\label{eq: two-view encoder}
\textbf{W}_A^i = \phi_\theta(\{\textbf{I}_A, \textbf{D}_A, \textbf{I}_B, \textbf{D}_B\}), \qquad
\textbf{W}_B^i = \phi_\theta(\{\textbf{I}_B, \textbf{D}_B, \textbf{I}_A, \textbf{D}_A\}),
\end{equation}
where $\textbf{W}_A^i$ and $\textbf{W}_B^i$ are the outputs of the two-view encoder at level $i$, $i \in {1,2,3,4}$, for frame $A$ and $B$ respectively and $\{ , \}$ is the concatenation operation. On each pyramid level, we extract the dense feature and uncertainty maps by feeding the two-view encoder outputs into the feature encoder branch and the uncertainty encoder branch:
\begin{equation}
\label{eq: feature encoder}
\textbf{F}_X^i = \phi_F(\textbf{W}_X^i), \qquad
\sigma_X^i = \phi_\sigma(\textbf{W}_X^i), 
\end{equation}
where $X \in {A, B}$. $i$ will be omitted later when we explain operation on the same pyramid level. Different from~\cite{Lv:etal:CVPR2019} which averages the output features map into one single channel, we maintain a same high-dimensional feature map at different pyramid levels. 
This choice is motivated by the hypothesis that higher dimensionality should lead to higher discriminative power of the features -- which we support in the experimental section.

\subsection{Probabilistic Feature-metric Residual for Pose Estimation}
In probabilistic estimation that assumes an underlying Gaussian distribution of the residuals, we equivalently minimise the weighted least squares, with the inverse covariance matrix acting as the weight.  
Given the dense feature and uncertainty maps on two views and an estimated pose $\mbs \xi_{AB}$, we propose a probabilistic feature-metric residual as an uncertainty-normalised feature difference: 
\begin{eqnarray}
\label{eq: probalistic-feature-residual}
\mbf r_{f} (\mbs \xi_{AB}) = \frac{\mbfbar r_{f} (\mbs \xi_{AB})}{\sigma_{f} (\mbs \xi_{AB})} = \frac{\mbf F_A[\pixel A (\mbs \xi_{AB})] - \mbf F_B[\pixel{B}(\mbs \xi_0)]}{\sqrt{\sigma_A^2[\pixel A (\mbs \xi_{AB})] + \sigma_B^2[\pixel{B}(\mbs \xi_0)]}},
\end{eqnarray}
where $\pixel A$ and $\pixel B$ are a pair of pixel correspondences on the two frames. $\mbfh u$ represents image pixel coordinates. $\pixel{B}(\mbs \xi_0)$ means $\pixel{B}$ is perturbed under zero transformation $\mbs \xi_0$. $\mbfbar r_{f}$ is the feature difference between the correspondences on the feature map and $\sigma_{f}$ is the joint uncertainty estimate for the correspondence that we obtain as a combination from the individual uncertainties. 
Note that this assumes isotropic uncertainty w.r.t.\ each feature dimension -- a simplification we chose (for speed) that may be revisited. Eq.~\ref{eq: probalistic-feature-residual} encourages the feature map from two different views to be as similar as possible while downweighs the features that the network is uncertain about from the either view with the predicted uncertainties. As shown in example Fig.~\ref{fig:teaser}, the trained features are robust to moderate lighting, reflection and view perspective variances and the trained uncertainties handle the uncertain features caused by the extreme lighting changes (lower right corner).
The dense correspondence lookup is implemented via warping from frame $B$ to frame $A$ through $\mbs \xi_{AB}$, which can be defined as:
\begin{eqnarray}
\label{eq: warping}
\pixel A (\mbs \xi_{AB}) = \pi(\T{A}{B}(\mbs \xi) \pi^{-1}(\pixel{B}, D_B[\pixel{B}])),
\end{eqnarray}
where $[.]$ represents the pixel lookup (including bilinear interpolation). $\pi$ and $\pi^{-1}$ denote the projection function to the image plane and the back-projection function to 3D (homogeneous) coordinates, respectively.
By inserting Eq.~\ref{eq: probalistic-feature-residual} into a Lucas-Kanade framework~\cite{Lucas:Kanade:IJCAI1981}, we formulate the pose estimation problem of an optimal pose $\mbs \xi^*$ as:
\begin{eqnarray}
\label{eq: featuremetric loss}
\mbs \xi^* = \argmin_{\mbs \xi} \displaystyle\frac{1}{2}\sum_{\pixel{B} \in \mathcal{U}}\mbf r_{f}^T(\mbs \xi) \mbf r_{f}(\mbs \xi),
\end{eqnarray}
i.e.\ summing all residuals over non-occluded pixels in $B$, $\mathcal{U}$, which can be iteratively solved by e.g.\ the Gauss-Newton method. To speed up the computation, we choose the inverse compositional formulation~\cite{Baker:Matthews:IJCV2004} that updates poses by applying the incremental pose on frame $B$. It allows for a more efficient computation of the feature-metric Jacobians. In each iteration, the pose is updated by $ \Delta \mbs \xi$ as: 
\begin{eqnarray}
\label{eq: pose update}
\mbs \xi_{k+1} =& \mbs \xi_{k} \circ \Delta \mbs \xi^{-1}, \\
\label{eq: pose solve}
\Delta \mbs \xi =& -(\mbf J_{f}^T \mbf J_{f})^{-1}(\mbf J_{f}^T \mbf r_{f}).
\end{eqnarray}
$\mbf J_{f}$ is the Jacobian of the probabilistic feature-metric residual $\mbs r_{f}$ w.r.t.\ the relative pose $\mbs \xi_{AB}$:
\begin{eqnarray}
\label{eq: Jac-ftr-def}
\mbf J_{f} = \frac{\partial \mbf r_{f}}{\partial \mbs \xi_{AB}} = 
-\left(\frac{\nabla \mbf F_B}{\sigma_{f} (\mbs \xi_{AB})} + \frac{\mbfbar r_{f} (\mbs \xi_{AB})\sigma_B \nabla \sigma_B }{\sigma^3_{f} (\mbs \xi_{AB})}\right)\frac{\partial \pixel{B}}{\partial \mbs \xi_0},
\end{eqnarray}
where $\nabla \mbf F_B$ and $\nabla \sigma_B$ are the gradients of the feature maps and uncertainty maps along the two pixel dimensions in frame $B$, respectively. Under this formulation, only the components of $\sigma_{f} (\mbs \xi)$ and $\mbfbar r_{f} (\mbs \xi)$ need to be re-evaluated in each iteration, which can be shared when computing the residuals in Eq.~\ref{eq: probalistic-feature-residual}. All the other components in Eq.~\ref{eq: Jac-ftr-def} can be pre-computed to speed up the computation.

\subsection{A Probabilistic Combination with ICP Residual}
As an uncertainty-driven residual, our proposed residual can be naturally combined with other residuals. For example, we can combine it with an ICP residual to add a more geometric constraint. The combined residual equation is:
\begin{eqnarray}
\label{eq: our-joint-residual}
\mbs \xi^* = \argmin_{\mbs \xi} \mbf r_{f}^T(\mbs \xi) \mbf r_{f}(\mbs \xi) +  w_{g} \mbf r_{g}^T(\mbs \xi) \mbs \Sigma^{-1}_{g} \mbf r_{g}(\mbs \xi),
\end{eqnarray}
where $\mbs r_{g}$ and $\mbs \Sigma_{g}$ are the ICP residual and uncertainty, respectively, and $w_{g}$ is the weight for ICP residual. The above equation can still be iteratively solved via the Gauss-Newton method. The detailed definitions of the ICP residual and Jacobian can be found in~\cite{Rusinkiewicz:Levoy:3DMIN2001}.
As there are no regularisation terms in Eq.~\ref{eq: probalistic-feature-residual}, our learned uncertainty is a scale-free parameter. When combining with other residuals of different magnitudes, we need to scale them properly before fine-tuning to bootstrap the training. The scale of ICP weight $w_{g}$ is chosen (as $w_{g}=0.01$) such that the individual Chi-square errors are of similar magnitude, after which the joint ICP/feature-metric training will scale the features and feature-metric uncertainties to be best balanced with the ICP.

\subsection{Coarse-to-fine Optimisation and Initialisation}
The cost functions in Eq.~\ref{eq: featuremetric loss} and \ref{eq: our-joint-residual} can be optimised in a coarse-to-fine way using damped Gauss-Newton optimisations, which is applied on 4 pyramid levels, with a fixed number of rolled-out iterations, i.e.\ 3, on each level. We added a small damping constant in Eq.~\ref{eq: pose solve} to prevent the matrix inversion to be ill-conditioned. Coarse-to-fine  optimisation methods 
are sensitive to coarse-level estimation, where the incorrect estimations will be propagated to finer levels and the iterative optimisation may get stuck in a wrong local minimum, especially in a wide-baseline setting. To tackle this issue, we train a pose network to bootstrap the optimisation by predicting an initial relative pose on the coarsest level, instead of using a conventional identity pose initialisation. To make the network compact, the concatenated outputs from the coarsest-level two-view encoder on the two frames serve as the inputs to our pose prediction network:
\begin{eqnarray}
\label{eq: posenet}
\mbs \xi_0 = \phi_T(\{\textbf{W}_A^1, \textbf{W}_B^1\}).
\end{eqnarray}
To account for the multi-modal information on the coarse level, the deep initial pose network outputs $K$ pose hypotheses, which are parameterised as 3 Euler angles and 3D translation vectors, and a respective confidence probability for each hypothesis. The final predicted pose is the weighted average of all hypotheses. 

\subsection{Training Setup}
The predicted initial pose and the estimated poses per pyramid level are compared to the ground truth pose and the resulting gradients in the optimisation are used for back-propagation to update all the learning weights. To balance influence of rotation vs.\ translation, we use the 3D End-Point-Error (EPE) as the training loss:
given the ground truth relative transformation $ \T{A}{B}(\mbs \xi)$ and the estimated/predicted pose $ \T{A}{B}(\mbs \xi_i)$, the loss is composed as:
\begin{equation}\label{eq:train_loss}
L = \frac{1}{|\mathcal V|}\sum_{i \in \mathcal I} \sum_{\generalThree{B}{v}{} \in \mathcal V} \norm{\T{A}{B}(\mbs \xi)\,\generalThree{B}{v}{} - \T{A}{B}(\mbs \xi_i)\,\generalThree{B}{v}{}}^{2}_2,
\end{equation}
where $\mathcal V$ is the set of backprojected 3D points $\generalThree{B}{v}{}$ in the frame $B$, $\mathcal I = \{0,1,2,3,4\}$ denotes the pyramid levels, $\mbs \xi_0$ is the predicted pose from the pose network and the other $\mbs \xi_i$ are the estimated poses at the final iteration of Gauss-Newton optimisations on the respective pyramid level. This formulation enables the network to learn both feature and uncertainty representations in an end-to-end fashion, without the need for a ground truth feature map or ground truth correspondences, and without requiring an explicit definition of the uncertainty model. We set the feature map channels to be 8. Note that the uncertainty is defined as a scalar value. We unroll the Gauss-Newton optimisation and train all the models together from scratch using ADAM~\cite{Kingma:Ba:ICLR2015} for 30 epochs, with a learning rate initialized at 0.0005 and reduced at epochs [5, 10, 20]. When combining the ICP residual, we do a further fine-tuning for 10 epochs.

\subsection{Implementation Details}
\label{subsec: implementation details}
\begin{figure}[t]
	\centering
	\includegraphics[width=\columnwidth]{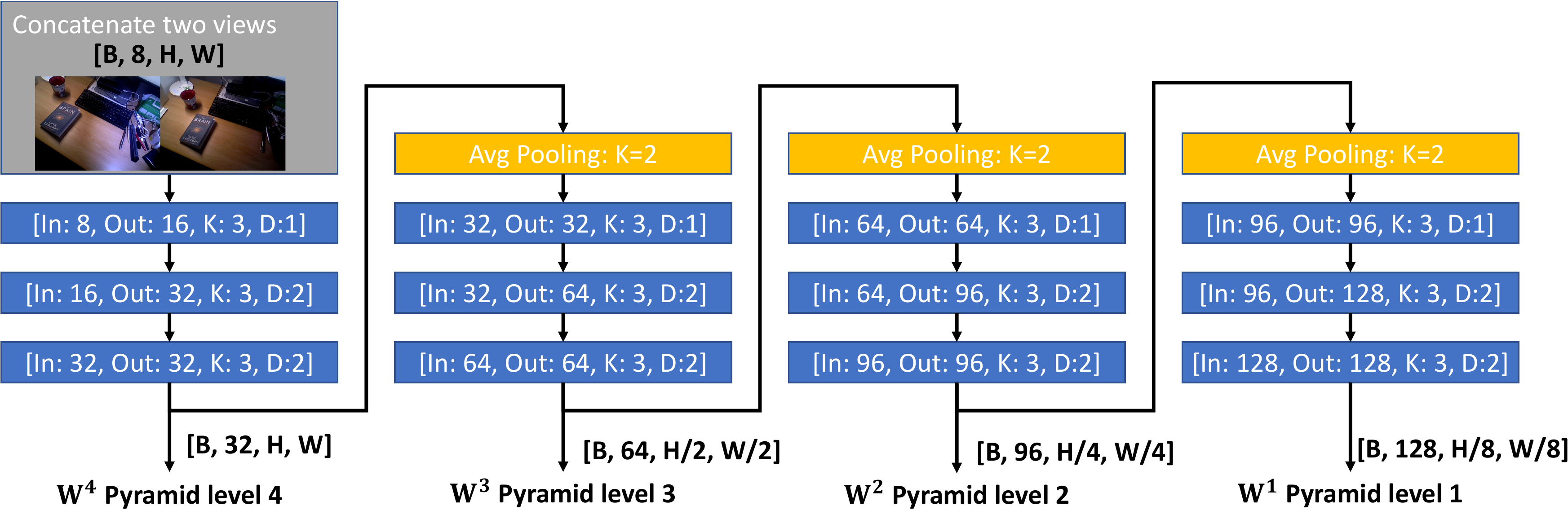}%
	\caption{The architecture of our two-view encoder. It is composed of basic convolutional blocks (blue) and average pooling operations (yellow). The basic convolutional block is grouped by a convolutional layer and followed by a BatchNorm layer, and a ELU layer. [In, Out, K, D] represents [Input channel, Output channel, Kernel size, Dilation] with stride always being 1. }
	\label{fig:Two-view encoder}
\end{figure}

Fig~\ref{fig:Two-view encoder} shows the architecture of our two-view encoder which takes the input from a pair of RGB-D images and extracts spatio-temporal correlation information from that. It is constructed into a 4-level pyramid architecture, where each level outputs a higher-dimension information. The architecture is modified from~\cite{Lv:etal:CVPR2019}, however, we do not perform an average operation to extract feature maps. Instead, we send the outputs to the feature encoder and the uncertainty encoder to estimate the feature and uncertainty maps.

\begin{figure}[t]
	\centering
	\includegraphics[width=\columnwidth]{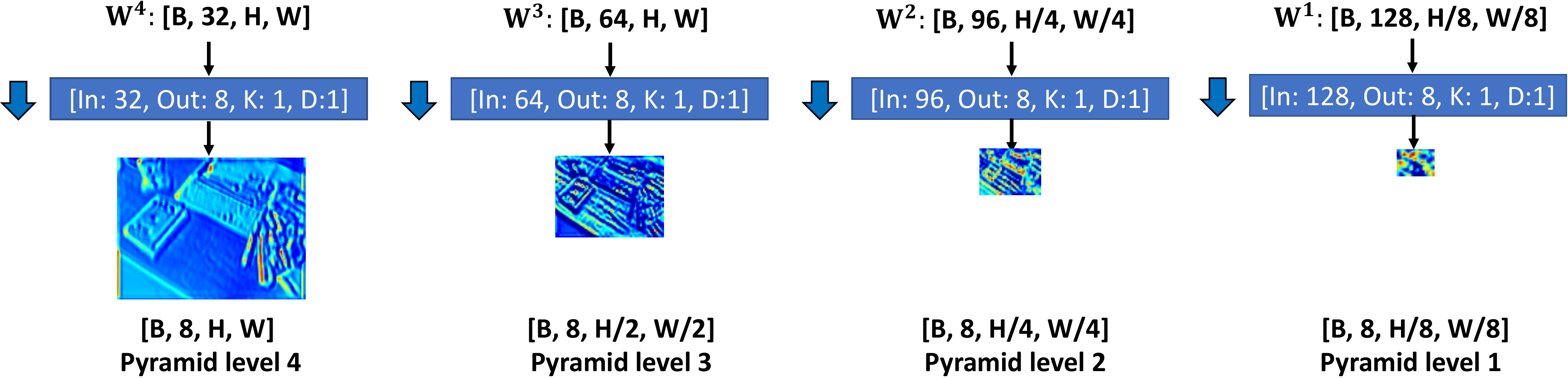}%
	\caption{The architecture of our feature encoder. On each pyramid level, it is a basic convolutional block that is group by a 1 by 1 convolutional layer, a BatchNorm layer, and a ELU layer. [In, Out, K, D] represents [Input channel, Output channel, Kernel size, Dilation] with stride always being 1.}
	\label{fig:feature encoder}
\end{figure}

Fig~\ref{fig:feature encoder} shows the architecture of our feature encoder on each pyramid level. It takes the input from the two-view encoder and predicts an 8-dimensional feature map.

\begin{figure}[t]
	\centering
	\includegraphics[width=\columnwidth]{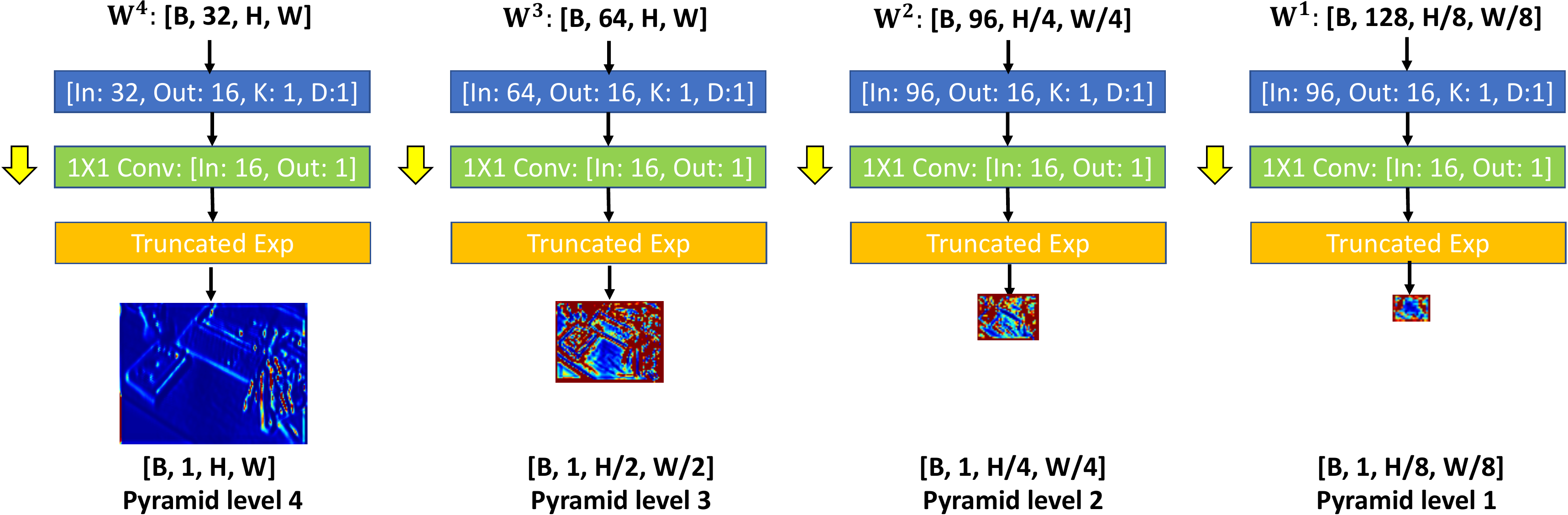}%
	\caption{The architecture of our uncertainty encoder. On each pyramid level, it is composed by a basic convolutional block, followed by a 1 by 1 convolutional layer and a truncated exponential operation. }
	\label{fig:uncertainty encoder}
\end{figure}

Fig~\ref{fig:uncertainty encoder} shows the architecture of our uncertainty encoder on each pyramid level. It takes the input from the two-view encoder and predicts a 1-dimensional uncertainty map. We assume the output from the 1 by 1 convolutional layer is a logarithmised uncertainty and we use the exponentiation operation to recover the true uncertainty. The output is truncated to avoid gradient explosion. 

\begin{figure}[t]
	\centering
	\includegraphics[width=0.9\columnwidth]{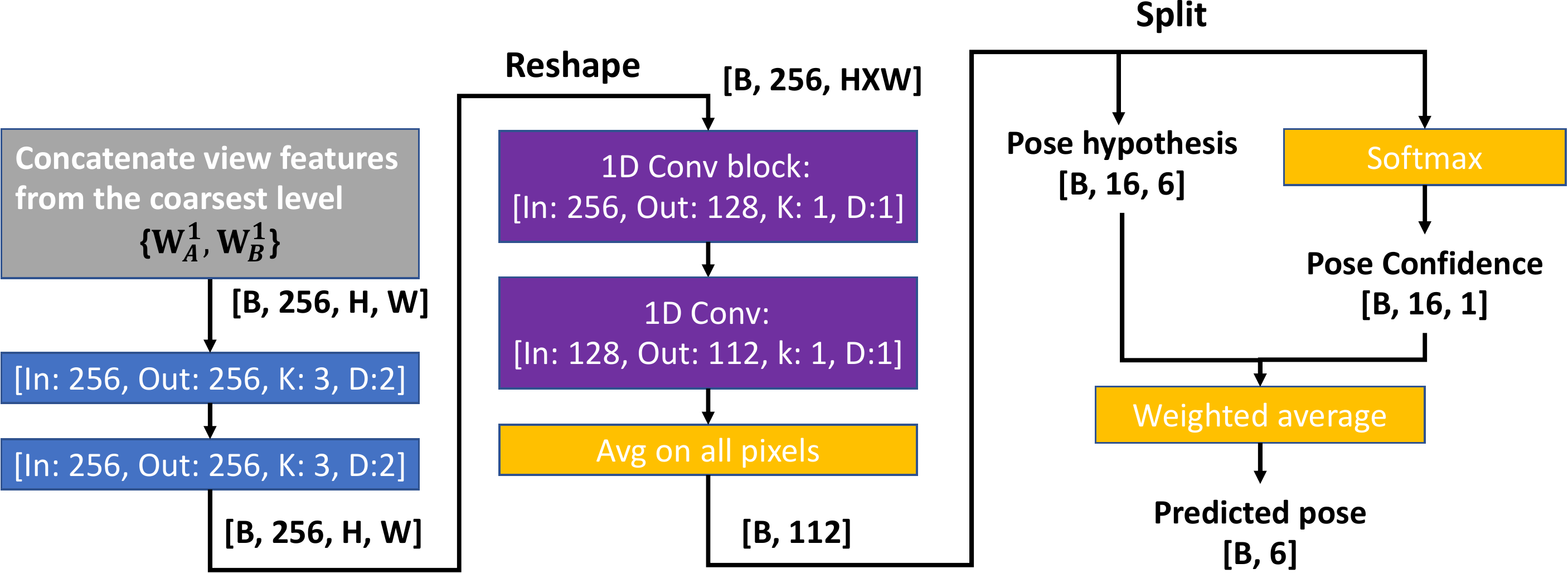}%
	\caption{The architecture of our pose network for initial pose prediction.}
	\label{fig:Posenet}
\end{figure}

Fig.~\ref{fig:Posenet} shows the architecture of our pose network to predict an initial pose on the coarsest level of the coarse-to-fine Gauss-Newton optimisation. It takes the input from a concatenation of the outputs of the two frames from the two-view encoder at the coarsest level. Similar to~\cite{Zhou:etal:ECCV2018}, the initial pose network also predicts multiple pose hypotheses and then fuse them together using their respective confidences. Here, we choose the hypotheses number to be 16. The pose is parameterised with 3 Euler angles and a 3-dimensional translation vector.

\section{Experiments}
\subsection{Quantitative Evaluation and Discussion}
We first evaluate our method on the \textbf{TUM RGB-D} SLAM dataset~\cite{Sturm:etal:IROS2012}. A natural extension is to apply it to 3D rigid object motion estimation, which we test on the \textbf{MovingObjects3D} dataset~\cite{Lv:etal:CVPR2019}. 

\textbf{DeepIC}~\cite{Lv:etal:CVPR2019} is chosen as our main baseline method, which learns dense feature map for pose optimisation. To have a fair comparison, we use the same experimental setting as theirs. We randomly subsampled frames $B$ at intervals \{1,2,4,8\} relative to frame $A$ from TUM RGB-D dataset~\cite{Sturm:etal:IROS2012} and \{1,2,4\} from MovingObjects3D dataset~\cite{Lv:etal:CVPR2019} to generate various motion magnitudes and tracking difficulties as the training pairs. A comparison to this approach would show the importance of the uncertainty prediction and the initial pose prediction in our proposed method. We further implemented \textbf{DeepIC+P}, an augmented variant of DeepIC~\cite{Lv:etal:CVPR2019}, with our pose prediction network to initialise their optimisation. The same number of iterations and pyramid levels are used as in our method. A comparison to it would further verify the contribution of our proposed probabilistic feature-metric loss. 

To have a comparison to deep pose prediction methods that directly predict a relative transformation from two views, we implemented a \textbf{coarse-to-fine PoseNet}, similar to the tracking part in DeepTAM~\cite{Zhou:etal:ECCV2018}. It is implemented on four pyramid levels for coarse-to-fine pose refinements, where the predicted pose from a coarser pyramid level would be used to bootstrap the prediction on a finer level. The network architecture is similar to our pose network but with different weights on different pyramid levels. A comparison to it would show a benefit of our learning-based optimisation approach for pose estimation. We further included the iterative refinement idea from~\cite{Li:etal:ECCV2018} to the coarse-to-fine PoseNet approach. The \textbf{iterative PoseNet} has 3 iteration refinements on each pyramid level. All the learning-based comparison approaches are trained end-to-end using the loss in Eq.~\ref{eq:train_loss}.

For the non-learning approaches, we compare our method to the pure geometric Point-to-Plane \textbf{ICP} method~\cite{Rusinkiewicz:Levoy:3DMIN2001}, which is essentially robust to illumination changes. We also include an \textbf{RGB-D VO} method~\cite{Steinbrucker:etal:ICCVW2011} in the camera motion evaluation. A comparison to these approaches would show benefits of learning-based approaches, in terms of larger convergence basin and better accuracy, even under challenging lighting conditions. 

To reveal the contribution of each component, we provide a detailed ablation study. We denote our system component, dense feature map, dense uncertainty map, deep initial pose prediction as F, U, P, respectively. We select the following settings. \textbf{Ours (F)}: We replace the uncertainty prediction with an identity uncertainty and disable the pose prediction with an identity pose initialisation. \textbf{Ours (F+P)}: We replace the uncertainty prediction with an identity uncertainty. \textbf{Ours (F+U)}: We disable the pose prediction and only use the proposed probabilistic feature-metric residual for alignment. \textbf{Ours (F+U+P)}: A full version of our probabilistic feature-metric tracking system. \textbf{Ours+ICP}: A combination of the probabilistic feature-metric and ICP residuals. All these combinations are implemented in coarse-to-fine optimisations, with the same number of iterations and pyramid levels as in the proposed method. 
\begin{table*}[tb]
\centering
\begin{tabular}{c|cccc}
\hline
\multirow{2}{*}{Method} & \multicolumn{4}{c}{3D EPE (cm) / RPE translation (cm) / RPE rotation (Deg)}                                                         \\ \cline{2-5} 
                        & \multicolumn{1}{c|}{KF 1}           & \multicolumn{1}{c|}{KF 2}           & \multicolumn{1}{c|}{KF 4}            & KF 8              \\ \hline
 ICP~\cite{Rusinkiewicz:Levoy:3DMIN2001}      & \multicolumn{1}{c|}{2.53/1.25/0.75} & \multicolumn{1}{c|}{5.12/2.57/1.47} & \multicolumn{1}{c|}{13.21/5.73/3.70} & 28.80/10.54/7.89 \\
RGB-D VO~\cite{Steinbrucker:etal:ICCVW2011}                & \multicolumn{1}{c|}{2.31/1.03/0.55} & \multicolumn{1}{c|}{4.38/2.81/1.39} & \multicolumn{1}{c|}{12.67/5.95/3.99} & 31.13/13.83/9.20 \\
Coarse-to-fine PoseNet~\cite{Zhou:etal:ECCV2018}   & \multicolumn{1}{c|}{1.88/1.91/0.80} & \multicolumn{1}{c|}{3.08/3.76/1.42} & \multicolumn{1}{c|}{5.82/7.30/2.76}  & 15.43/13.16/5.73  \\ 
Iterative PoseNet~\cite{Zhou:etal:ECCV2018, Li:etal:ECCV2018}   & \multicolumn{1}{c|}{1.76/1.86/0.84} & \multicolumn{1}{c|}{2.70/3.61/1.53} & \multicolumn{1}{c|}{4.75/7.28/2.73}  & 12.74/13.12/5.23  \\ 
DeepIC~\cite{Lv:etal:CVPR2019}                  & \multicolumn{1}{c|}{1.31/0.69/0.45} & \multicolumn{1}{c|}{1.57/1.14/0.63} & \multicolumn{1}{c|}{2.53/2.09/1.10}  & 11.03/5.88/3.76  \\ 
DeepIC+P, adapted from~\cite{Lv:etal:CVPR2019}       & \multicolumn{1}{c|}{1.26/0.69/0.44} & \multicolumn{1}{c|}{1.46/1.13/0.60} & \multicolumn{1}{c|}{2.32/2.68/1.10}  & 8.20/5.06/3.73  \\ 
\hline
Ours (F)               & \multicolumn{1}{c|}{1.25/0.67/0.44} & \multicolumn{1}{c|}{1.49/1.14/0.60} & \multicolumn{1}{c|}{2.50/2.78/1.14}  & 11.70/12.20/4.37 \\

Ours (F+P)             & \multicolumn{1}{c|}{1.24/0.65/0.44} & \multicolumn{1}{c|}{1.42/1.04/0.57} & \multicolumn{1}{c|}{2.04/2.06/0.81}  & 7.35/6.71/2.89 \\

Ours (F+U)             & \multicolumn{1}{c|}{1.23/0.58/0.41} & \multicolumn{1}{c|}{1.40/0.86/0.50} & \multicolumn{1}{c|}{2.33/1.99/0.87}  & 13.24/12.92/4.59 \\
Ours (F+U+P)        & \multicolumn{1}{c|}{1.23/0.57/\textbf{0.40}} & \multicolumn{1}{c|}{1.38/0.80/0.48} & \multicolumn{1}{c|}{\textbf{1.71}/\textbf{1.22}/\textbf{0.64}}  & 5.48/4.89/2.12   \\
Ours+ICP      & \multicolumn{1}{c|}{\textbf{1.22}/\textbf{0.54}/\textbf{0.40}} & \multicolumn{1}{c|}{\textbf{1.33}/\textbf{0.76}/\textbf{0.47}} & \multicolumn{1}{c|}{1.78/1.26/0.66}  & \textbf{4.82}/\textbf{4.57}/\textbf{2.00}   \\ \hline
\end{tabular}
\vspace{0.5em}
\caption{Results on our test split in TUM RGB-D Dataset. KF denotes the frame intervals. }
\vspace{-0.5em}
\label{tab:tum_quantitative}
\end{table*}
The evaluation metrics are the 3D EPE loss in Eq.~\ref{eq:train_loss} and the relative pose error (RPE) metrics defined in TUM RGB-D dataset~\cite{Sturm:etal:IROS2012}. 

\textbf{TUM RGB-D Dataset:}
We use the same setting as DeepIC~\cite{Lv:etal:CVPR2019}, where sequences `fr1/360', `fr1/desk', `fr2/360', and `fr2/pioneer360' are used for testing and the remaining sequences are split into training (first 95\% of each sequence) and validation (last 5\%). Images are transformed to a resolution of 160$\times$120, with depth values outside of 0.5m to 5.0m being ignored.

Table~\ref{tab:tum_quantitative} summarises the results on the TUM RGB-D dataset. Our method outperforms all the other state-of-the-art learning-based approaches, as well as the non-learning RGB-D VO, and ICP methods, from small baselines to large baselines. Compared with all ablation variants, our full version (F+U+P) achieves the best performance. The addition of uncertainty estimation complements the high-dimensional feature-metric alignment to improve the tracking accuracy. The predicted initial pose further improves the accuracy by bringing the estimation close the correct minimum, especially in the large motion scenarios. After fine-tuning the probabilistic combination with ICP loss, it can be seen that the performance is further improved in most cases (except KF 4 where the performance drops a bit), showing the validity of the probabilistic combination. 

We have further developed a prototype visual odometry system, where the camera pose is estimated by our proposed method. 
Despite being a pure frame-to-frame tracking system without components of keyframing and loop closure optimisations, drift caused by incremental misalignment qualitatively remains small. The qualitative results can be found in the supplementary video.

\textbf{MovingObjects3D Dataset:}
MovingObjects3D dataset contains 6 different catogories of objects moving in front of the camera under various illumination changes. We follow the dataset setting, where the categories of `boat' and `motorbike' are used as the testing set and the other categories are split into training (first 95\% sequences of each category) and validation (last 5\%), to test tracking performance for unseen objects. For the non-learning-based ICP~\cite{Rusinkiewicz:Levoy:3DMIN2001} approach, we provide ground truth object masks for them to test their optimal performances. For the learning-based approaches, we reply on those systems to distinguish the object motion from the background, given the ground truth object and camera motions. Table~\ref{tab:objects_quantitative} reports the results, which again show the superior performance of our method and confirm the contribution of each proposed component. 
\begin{table*}[htb]
\centering
\begin{tabular}{c|ccc}
\hline
\multirow{2}{*}{Method} & \multicolumn{3}{c}{3D EPE (cm) / RPE translation (cm) / RPE rotation (Deg)}                     \\ \cline{2-4} 
                        & \multicolumn{1}{c|}{KF 1}            & \multicolumn{1}{c|}{KF 2}            & KF 4              \\ \hline
ICP~\cite{Rusinkiewicz:Levoy:3DMIN2001}      & \multicolumn{1}{c|}{3.31/9.75/\textbf{2.74}}  & \multicolumn{1}{c|}{9.63/19.72/8.31} & \multicolumn{1}{c}{19.98/41.40/16.64}                    \\
Coarse-to-fine PoseNet~\cite{Zhou:etal:ECCV2018}   & \multicolumn{1}{c|}{2.62/10.10/4.14}  & \multicolumn{1}{c|}{5.01/20.19/8.29} & \multicolumn{1}{c}{9.63/38.96/16.02} \\ 
Iterative PoseNet~\cite{Zhou:etal:ECCV2018, Li:etal:ECCV2018}   & \multicolumn{1}{c|}{2.55/10.08/4.14}  & \multicolumn{1}{c|}{4.96/20.16/8.28} & \multicolumn{1}{c}{9.60/38.91/16.00} \\ 
DeepIC~\cite{Lv:etal:CVPR2019}                  & \multicolumn{1}{c|}{2.91/9.73/3.74}  & \multicolumn{1}{c|}{5.94/19.60/7.41} & \multicolumn{1}{c}{12.96/38.39/14.71} \\ 
DeepIC+P, adapted from~\cite{Lv:etal:CVPR2019}   & \multicolumn{1}{c|}{2.66/9.78/3.76}  & \multicolumn{1}{c|}{5.14/19.72/7.67} & \multicolumn{1}{c}{9.90/38.50/15.17} \\ 
\hline
Ours (F)               & \multicolumn{1}{c|}{2.52/9.34/3.57} & \multicolumn{1}{c|}{5.04/18.90/7.26} & \multicolumn{1}{c}{10.49/37.19/14.39} \\
Ours (F+P)             & \multicolumn{1}{c|}{2.64/9.59/3.64}  & \multicolumn{1}{c|}{5.14/19.42/7.43} & \multicolumn{1}{c}{9.97/37.01/14.32}  \\
Ours (F+U)             & \multicolumn{1}{c|}{2.20/8.62/3.43}  & \multicolumn{1}{c|}{4.53/17.90/7.19} & \multicolumn{1}{c}{9.86/36.18/14.50}  \\
Ours (F+U+P)        & \multicolumn{1}{c|}{2.17/8.44/3.22}  & \multicolumn{1}{c|}{4.47/17.86/6.91} & \multicolumn{1}{c}{9.26/36.443/14.22} \\
Ours+ICP      & \multicolumn{1}{c|}{\textbf{1.93}/\textbf{7.84}/2.93}        & \multicolumn{1}{c|}{\textbf{4.12}/\textbf{16.94}/\textbf{6.29}}  & \multicolumn{1}{c}{\textbf{8.93}/\textbf{35.39}/\textbf{13.14}}                  \\ \hline
\end{tabular}
\vspace{0.5em}
\caption{Results on our test split of MovingObjects3D Dataset.}
\vspace{-0.5em}
\label{tab:objects_quantitative}
\end{table*}

\begin{figure}[t]
\centering 
	\subfloat[][Frame A] {\includegraphics[width=0.24\columnwidth]{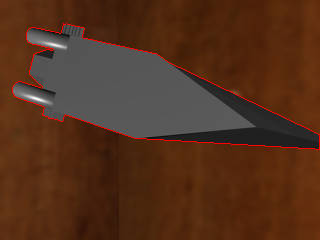} }
	\subfloat[][Frame B] {\includegraphics[width=0.24\columnwidth]{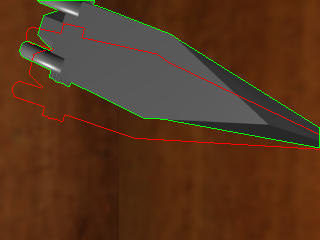} } 
	\subfloat[][Ours] {\includegraphics[width=0.24\columnwidth]{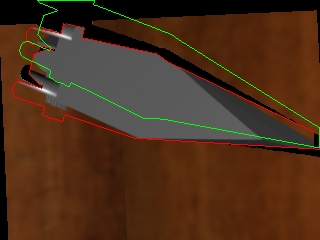} } 
	\subfloat[][Ours+ICP] {\includegraphics[width=0.24\columnwidth]{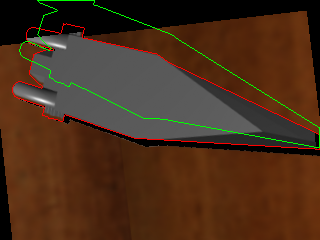} } \\
	\caption{Qualitative results on MovingObjects3D dataset. Object motion between the frame A and frame B is estimated using our proposed method (c) and a further combination with ICP (d). The object is warped from frame A to B using the estimated motion for visualization. The ground truth object boundaries in $A$ and $B$ are colored in red and color, respectively. Black regions in the warped image are caused by occlusion.}
	\label{fig:example_obj}
\end{figure}

Figure \ref{fig:example_obj} visualises our tracking result on the test split of MovingObjects3D dataset. As can be seen, our proposed method can provide a good alignment for objects under large motion and lighting changes.  A combination with ICP can provide a further refinement in the pose estimation.  

\textbf{Ablation Study on the Choice of Channel Dimension:}
As examined in~\cite{Czarnowski:etal:ICCVW2017}, multi-dimensional feature map from network can improve tracking robustness. In Table~\ref{tab:tum_quantitative} and~\ref{tab:objects_quantitative}, \textbf{Ours (F)}, with higher-dimension features, outperforms~\cite{Lv:etal:CVPR2019} in most cases, even without uncertainty or pose predictions. On the other hand, a higher dimension of feature maps usually bring a higher computational cost. In this part, we experimentally evaluate the effect of the channel dimension of the feature map and the uncertainty map. We fix the uncertainty channel to be 1 when we vary the feature channels and fix the feature channel to be 8 when we vary the uncertainty channels between 1 and the same feature channel, i.e.\ 8. Table~\ref{tab: abla-channel-dimension} summarises accuracy and inference time on the TUM RGB-D dataset~\cite{Sturm:etal:IROS2012}. Note that the accuracy increases when we increase the channel dimension of feature map, albeit with diminishing gains at dimensions higher than 8. When we increase the channel dimension of the uncertainty map, the accuracy very slightly increases for small baselines and slightly decreases for large baselines, validating the original choice of scalar uncertainty prediction.

In addition to accuracy, the increase of channel dimension in either feature or uncertainty map dimension would increase the GPU memory usage and reduce the inference speed. As a compromise of all these factors, we choose the feature dimension to be 8 and the uncertainty dimension to be 1 in all our other experiments. 

\begin{table*}[htb]
	\centering
	\begin{tabular}{cc|ccccc}
		\hline
		\multirow{2}{*}{Map} & \multirow{2}{*}{C} & \multicolumn{4}{c}{3D EPE (cm) / RPE translation (cm) / RPE rotation (Deg)}                                                      & \multirow{2}{*}{\begin{tabular}[c]{@{}c@{}} Time \\ (ms)\end{tabular}} \\ \cline{3-6}
		&                             & \multicolumn{1}{c|}{KF 1}           & \multicolumn{1}{c|}{KF 2}           & \multicolumn{1}{c|}{KF 4}           & KF8            &                                                                               \\ \hline
		\multirow{4}{*}{\begin{tabular}[c]{@{}c@{}} F \\ U=1 \end{tabular}}     & 1                       & \multicolumn{1}{c|}{1.23/0.58/0.41} & \multicolumn{1}{c|}{1.37/0.83/0.50} & \multicolumn{1}{c|}{1.86/1.48/0.74} & 8.15/6.09/2.93 & 5.41\\
		& 3                       & \multicolumn{1}{c|}{1.23/\textbf{0.57}/\textbf{0.40}} & \multicolumn{1}{c|}{1.36/\textbf{0.78}/\textbf{0.48}} & \multicolumn{1}{c|}{1.72/1.24/0.64} & 5.92/5.05/2.20 & 6.25 \\
		& 8                       & \multicolumn{1}{c|}{1.23/\textbf{0.57}/\textbf{0.40}} & \multicolumn{1}{c|}{1.38/0.80/\textbf{0.48}} & \multicolumn{1}{c|}{1.71/1.22/0.64} & \textbf{5.48}/\textbf{4.89}/\textbf{2.12} & 7.29 \\
		& 16                      & \multicolumn{1}{c|}{\textbf{1.22}/\textbf{0.57}/\textbf{0.40}} & \multicolumn{1}{c|}{\textbf{1.35}/\textbf{0.78}/\textbf{0.48}} & \multicolumn{1}{c|}{\textbf{1.66}/\textbf{1.21}/\textbf{0.62}} & 5.72/4.94/2.22 & 11.67 \\ \hline
		\multirow{2}{*}{\begin{tabular}[c]{@{}c@{}} U \\ F=8 \end{tabular}} & 1                       & \multicolumn{1}{c|}{1.23/0.57/\textbf{0.40}} & \multicolumn{1}{c|}{1.38/0.80/\textbf{0.48}} & \multicolumn{1}{c|}{\textbf{1.71}/\textbf{1.22}/\textbf{0.64}} & \textbf{5.48}/\textbf{4.89}/\textbf{2.12} & 7.29 \\
		& 8                       & \multicolumn{1}{c|}{\textbf{1.22}/\textbf{0.55}/\textbf{0.40}} & \multicolumn{1}{c|}{\textbf{1.37}/\textbf{0.79}/0.49} & \multicolumn{1}{c|}{1.74/1.35/0.67} & 6.15/5.58/2.38 & 9.13 \\ \hline
	\end{tabular}
	\vspace{0.5em}
	\caption{Ablation study of the channel dimension effect on our test split in TUM RGB-D. F, U, C abbreviate the feature map, uncertainty, and the channel dimension. Time is the average inference time for a pair of input RGB-D images (size 160$\times$120). }
	\vspace{-0.5em}
	\label{tab: abla-channel-dimension}
\end{table*}

\textbf{Model Size and Computation Time:}
Our system implemented in PyTorch has 1.83M learnable parameters. The average forward inference time for a pair of RGB-D image in the resolution of 160$\times$120 on a GTX 1080 platform is 7.29ms. After integrating ICP, it is 9.84ms (i.e. +35\%) on the same platform.

We also studied the effect of the input image resolution. With increased resolution (256$\times$192), accuracy slightly improves on the small baselines, i.e.\ KF 1 and 2, however, slightly deteriorates on KF 4 and 8 while the computation increases to 15.29ms (i.e.\ +111\%). Therefore, we set 160$\times$120 as main setting for training and testing.

\begin{figure}[tb]
\centering 
        \includegraphics[width=\columnwidth]{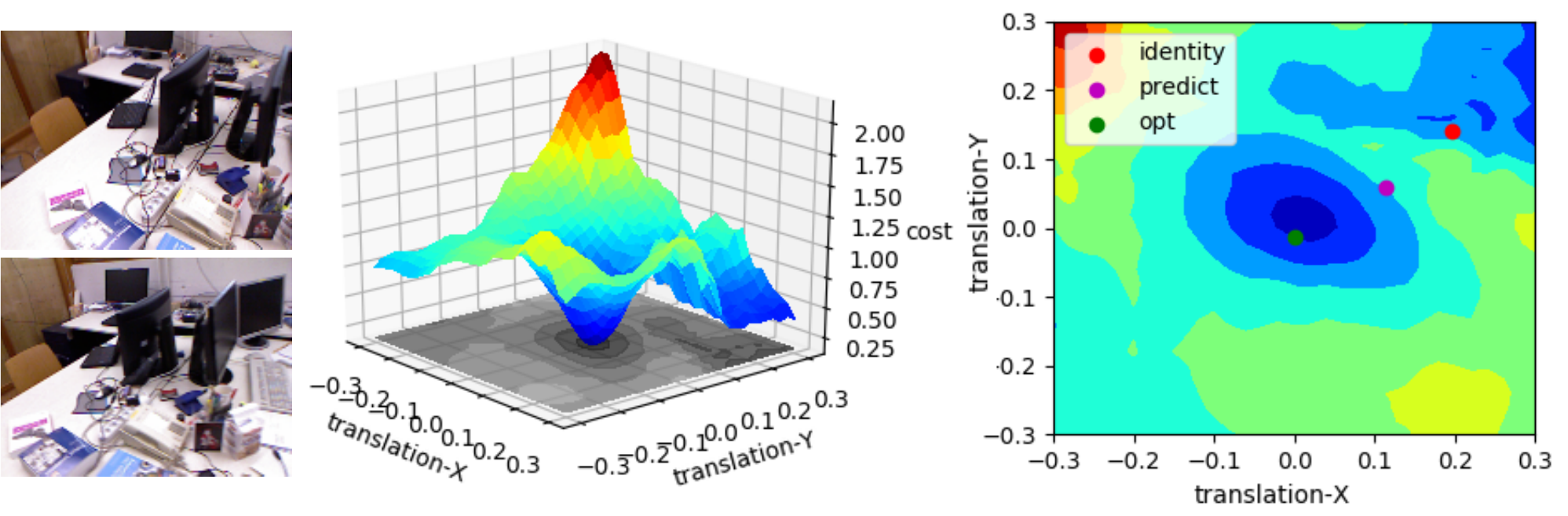}
    \caption{Visualisation of cost landscape of x and y translation for the feature-metric loss on the coarsest level. From left to right: input, cost landscape 3D, and 2D projection of cost landscape.}
	\label{fig:convergence-basin}
\end{figure}
\begin{figure}[htb]
\centering 
    \includegraphics[width=0.8\columnwidth]{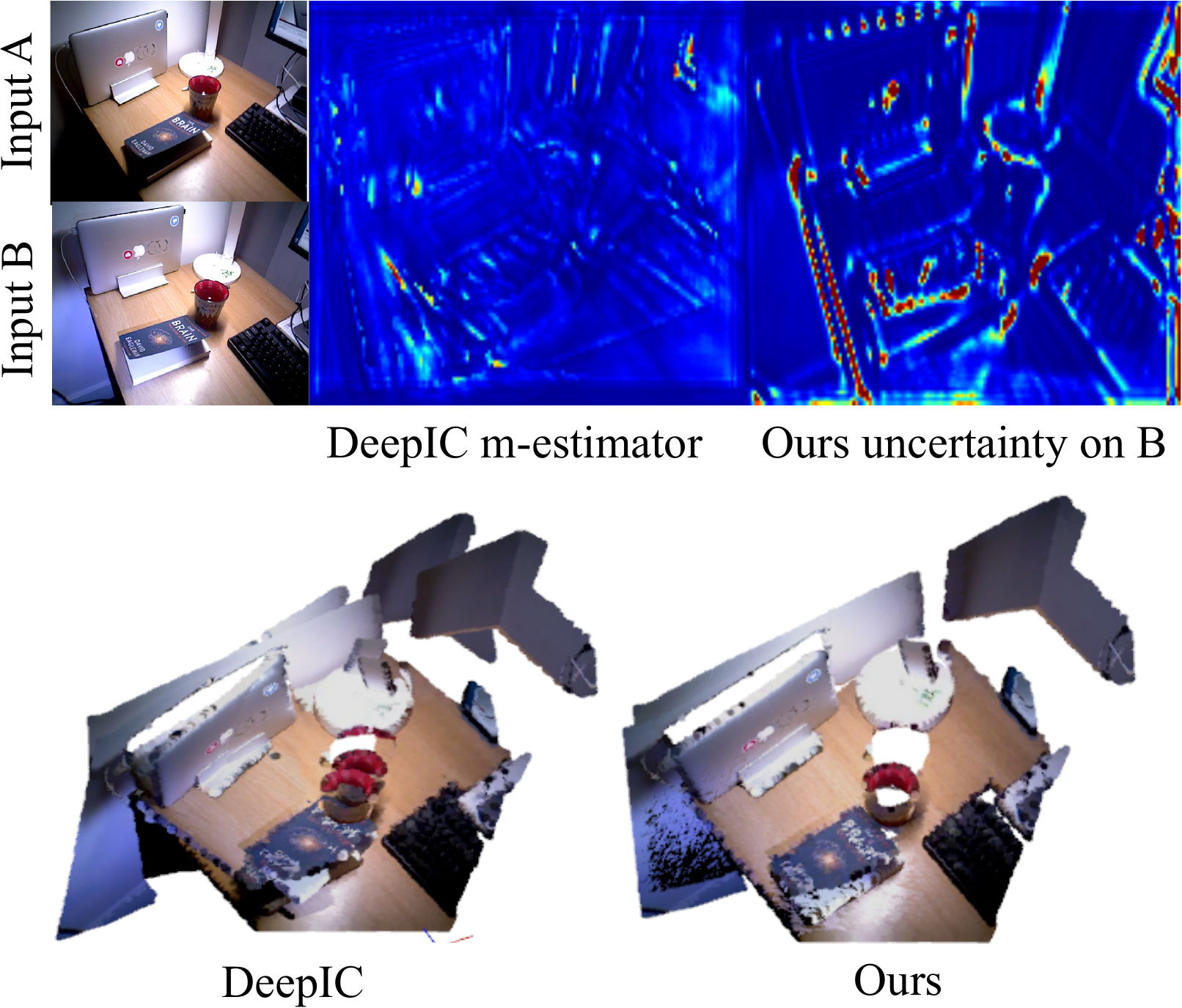}
    \caption{Qualitative evaluation in challenging lighting. Notice our uncertainty estimation is more sensitive to the lighting changes than the learned m-estimator in DeepIC (higher value is in red and lower value is in blue).}
	\label{fig:qua_mac_align}
\end{figure}
\subsection{Qualitative Evaluation and Discussions}
\textbf{Convergence Basin:~}
To analyse the effect of the initial pose prediction in our system, we perform a cost landscape visualisation experiment. Since $\mbs \xi$ is a 6D vector, it is computationally infeasible to sample cost on all possible pose components and also difficult to visualise the 6D cost landscape. Therefore, we choose to fix the rotation and z-translation components and only sample the pose combinations at the x and y translations around the ground truth pose. Fig.~\ref{fig:convergence-basin} shows one example on our test split from the TUM RGB-D dataset using the an interval of 8 frames. It can be seen that our pose prediction network brings the estimation into the convergence basin near the global minimum otherwise the conventional identity pose initialisation would lead the optimisation to a wrong local minimum.
\\\textbf{Challenging Illuminations:~}
Uncertainty prediction is significant for deploying neural network on robotic applications.
DeepIC~\cite{Lv:etal:CVPR2019} proposed a learned robust cost function m-estimator to downweigh the residual outliers. To evaluate our learned uncertainty and also to compare to DeepIC's learned m-estimator, we captured sequences using an RGB-D camera while we were waving a flashlight to create illumination changes. The collected sequences contain both local and global lighting, reflection, and shading variances across the images.
Since we don't have ground truth poses on these frames, we warp the point cloud from one frame to another using the estimated transformation between them and visualise the 3D pointcloud alignment of the two views. We test it using the weights trained from the TUM RGB-D dataset without fine-tuning. Fig.~\ref{fig:qua_mac_align} shows one example. It can be seen that our method provides more robust pose estimation under those lighting changes. This is partially because our estimated uncertainty can more reliablely capture illumination variance, e.g.\ on the book and desk surface, than DeepIC's m-estimator. Please refer to the supplementary video for more results and details. 

\section{Conclusion}

We presented a deep probabilistic feature-metric two-frame RGB-D tracking method by combining the power of deep learning for feature learning, uncertainty estimation and pose prediction in a learning-based optimisation framework. It enables our method compact and to outperform the state of the art methods on camera motion and rigid object motion estimation benchmarks. Challenging experiments have shown an accurate and robust performance under large motion and strong lighting change scenarios, which is significant and currently lacking, in real-world robotic applications. We further showcased how our proposed residual can easily be combined with commonly used ICP residual in practice. Continuing from here, we would like to explore how to better combine the probabilistic feature-metric residuals with other residuals. 
Also, we aim to apply our tracking method to full dense SLAM systems, including object-level and dynamic SLAM systems.

\section*{ACKNOWLEDGMENTS}
We wish to thank Shuaifeng Zhi, Jan Czarnowski and Tristan Laidlow for fruitful discussions and the anonymous reviewers for their comments and
advice.

\bibliographystyle{IEEEtran}
\bibliography{robotvision}

\end{document}